\documentclass[11pt,letterpaper]{article}
\usepackage{emnlp2017}
\usepackage{times}
\usepackage{latexsym}

\usepackage{url}
\usepackage{graphicx}
\usepackage{amsmath}
\usepackage{amsfonts}
\usepackage{algorithm}
\usepackage{algorithmic}

\usepackage{float}
\usepackage{multirow}
\usepackage{verbatim}
\usepackage{caption}
\usepackage{subcaption}
\usepackage{array}
\usepackage[flushleft]{threeparttable}
\usepackage{soul}

\usepackage[toc,page]{appendix}
\usepackage{booktabs}
\usepackage{float}

\emnlpfinalcopy

\def\emnlppaperid{25}

\title{Reader-Aware Multi-Document Summarization: An Enhanced Model and The First Dataset\Thanks{The work described in this paper is supported by a grant from the Grant Council of the Hong Kong Special Administrative Region, China (Project Code: 14203414).}}

\author{ Piji Li$^{\dag}$ \ \ Lidong Bing$^{\ddag}$ \ \  Wai Lam$^{\dag}$\\
	$^{\dag}$Department of Systems Engineering and Engineering Management,\\
	The Chinese University of Hong Kong\\
	$^{\ddag}$AI Lab, Tencent Inc., Shenzhen, China\\
	{\tt  $^{\dag}$\{pjli, wlam\}@se.cuhk.edu.hk, $^{\ddag}$lyndonbing@tencent.com}}

\date{}

\begin{document}

\maketitle

\begin{abstract}
  We investigate the problem of reader-aware multi-document summarization (RA-MDS) and introduce a new dataset for this problem. To tackle RA-MDS, we extend a variational auto-encodes (VAEs) based MDS framework by jointly considering news documents and reader comments. To conduct evaluation for summarization performance, we prepare a new dataset. We describe the methods for data collection, aspect annotation, and summary writing as well as scrutinizing by experts. Experimental results show that reader comments can improve the summarization performance, which also demonstrates the usefulness of the proposed dataset. The annotated dataset for RA-MDS is available online\footnote{\url{http://www.se.cuhk.edu.hk/\~textmine/dataset/ra-mds/}}.
\end{abstract}

\section{Introduction}

The goal of multi-document summarization (MDS) is to automatically generate a brief, well-organized summary for a topic which describes an event with a set of documents from different sources. \cite{goldstein2000multi,erkan2004lexrank,wan2007manifold,nenkova2012survey,min2012exploiting,lidong15absmds,li2017salience}.
In the typical setting of MDS, the input is a set of news documents about the same topic.
The output summary is a piece of short text document containing several sentences, generated only based on the input original documents.

\begin{figure}[!t]
	\centering
	\includegraphics[width=1\columnwidth]{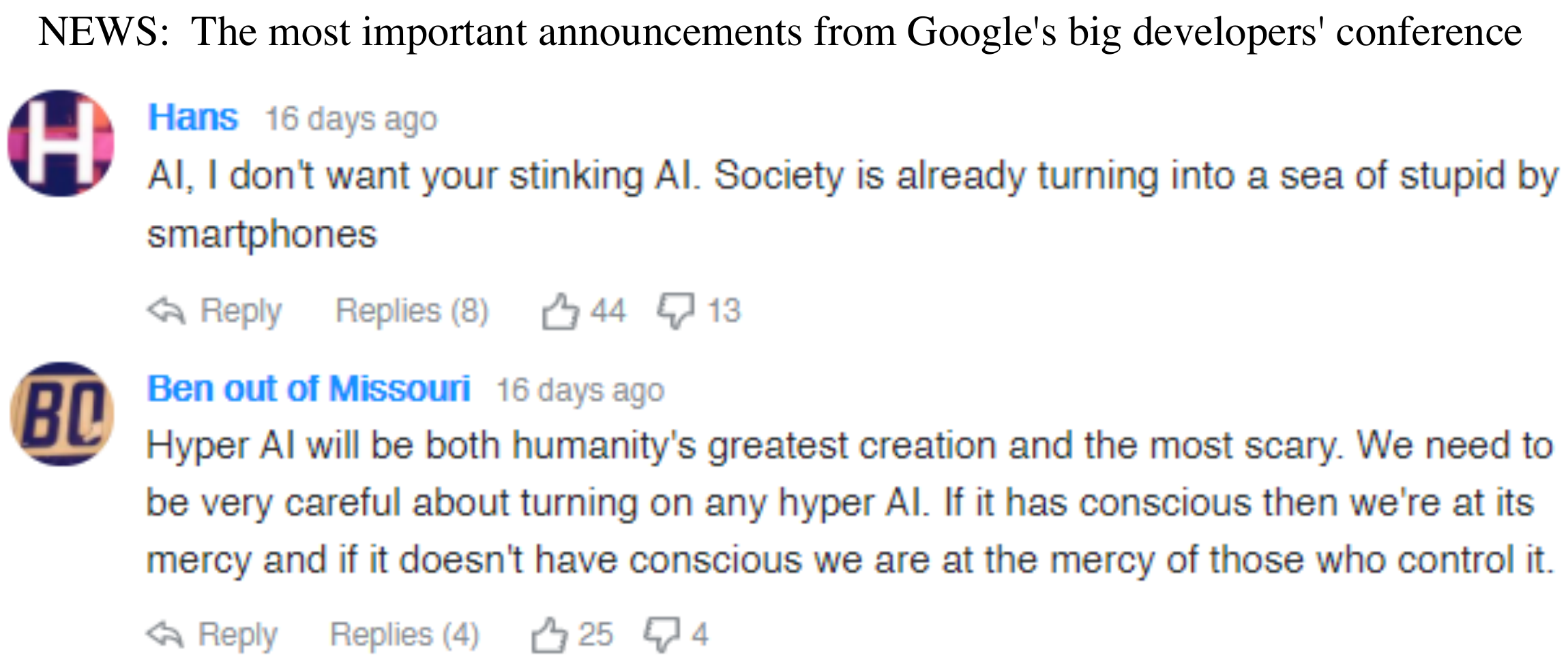}
	\caption{
		Reader comments of the news  ``The most important announcements from Google's big developers' conference (May, 2017)''.
	}
	\label{fig:front}
\end{figure}

With the development of social media and mobile equipments, more and more user generated content is available. Figure~\ref{fig:front} is a snapshot of reader comments under the news report ``The most important announcements from Google's big developers' conference''\footnote{https://goo.gl/DdU0vL}. The content of the original news report talks about some new products based on AI techniques. The news report generally conveys an enthusiastic tone. However, while some readers share similar enthusiasms, some others express their worries about new products and technologies and these comments can also reflect their interests which may not be very salient in the original news reports.
Unfortunately, existing MDS approaches cannot handle this issue.
We investigate this problem known as reader-aware multi-document summarization (RA-MDS). Under the RA-MDS setting, one should jointly consider news documents and reader comments when generating the summaries.

One challenge of the RA-MDS problem is how to conduct salience
estimation by jointly considering the focus of news reports and the reader interests revealed by comments. Meanwhile, the model should be insensitive to the availability of diverse aspects of reader comments.
Another challenge is that reader comments are very noisy, not fully grammatical and often expressed in informal expressions.
Some previous works explore the effect of comments or social contexts in
single document summarization such as blog summarization~\cite{Hu:2008:CDS:1390334.1390385,Yang:2011:SCS:2009916.2009954}.
However, the problem setting of RA-MDS is more challenging because the considered comments are about an event which is described by multiple documents spanning a time period. Another challenge is that reader comments are very diverse and noisy.
Recently, \citet{li2015reader} employed a sparse coding based framework for RA-MDS jointly considering news documents and reader comments via an unsupervised data reconstruction strategy. However, they only used the bag-of-words method to represent texts, which cannot capture the complex relationship between documents and comments.

Recently, \citet{li2017salience} proposed a sentence salience estimation framework known as \textit{VAESum} based on a neural generative model called Variational Auto-Encoders (VAEs) \cite{kingma2013auto,rezende2014stochastic}.
During our investigation, we find that the Gaussian based VAEs have a strong ability to capture the salience information and filter the noise from texts. Intuitively, if we feed both the news sentences and the comment sentences into the VAEs, commonly existed latent aspect information from both of them will be enhanced and become salient.
Inspired by this consideration, to address the sentence salience estimation problem for RA-MDS by jointly considering news documents and reader comments, we extend the VAESum framework by training the news sentence latent model and the comment sentence latent model simultaneously by sharing the neural parameters.
After estimating the sentence salience, we employ a phrase based compressive unified optimization framework to generate a final summary.

There is a lack of high-quality dataset suitable for RA-MDS.
Existing datasets from DUC\footnote{http://duc.nist.gov/} and TAC\footnote{http://tac.nist.gov/} are not appropriate.
Therefore, we introduce a new dataset for RA-MDS.
We employed some experts to conduct the tasks of data collection, aspect annotation, and summary writing as well as scrutinizing.
To our best knowledge, this is the first dataset for RA-MDS.

Our contributions are as follows:
(1) We investigate the RA-MDS problem and introduce a new dataset for the problem of RA-MDS. To our best knowledge, it is the first dataset for RA-MDS.
(2) To tackle the RA-MDS, we extend a VAEs-based MDS framework by jointly considering news documents and reader comments.
(3) Experimental results show that reader comments can improve the summarization performance, which also demonstrates the usefulness of the dataset.

\section{Framework}
\subsection{Overview}
\begin{figure*}[!t]
	\centering
	\includegraphics[width=1.7\columnwidth]{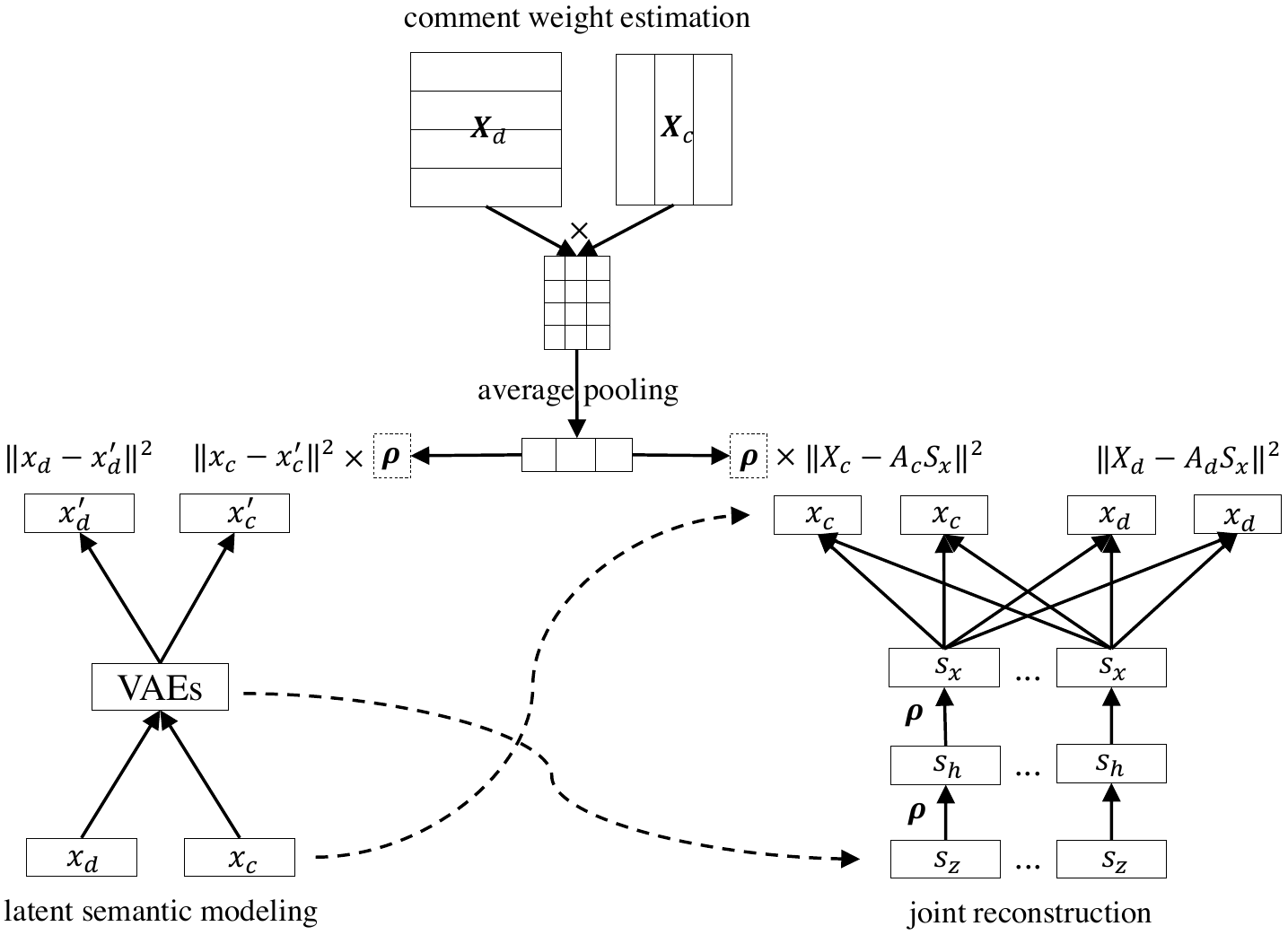}
	\caption{\label{fig:framework}
		Our proposed framework. \textbf{Left}:
		Latent semantic modeling via variation auto-encoders for news sentence $\mathbf{x}_d$ and comment sentence $\mathbf{x}_c$. \textbf{Middle}: Comment sentence weight estimation.  \textbf{Right}:
		Salience estimation by a joint data reconstruction method. $\mathbf{A}_d$ is a news reconstruction coefficient matrix which contains the news sentence salience information.
	}
\end{figure*}
As shown in Figure~\ref{fig:framework}, our reader-aware news sentence salience framework has three main components: (1) latent semantic modeling; (2) comment weight estimation; (3) joint reconstruction.
Consider a dataset $X_d$ and $X_c$ consisting of $n_d$ news sentences and $n_c$ comment sentences respectively from all the documents in a topic (event), represented by bag-of-words vectors.
Our proposed news sentence salience estimation framework is extended from VAESum~\cite{li2017salience}, which can jointly consider news documents and reader comments. One extension is that, in order to absorb more useful information and filter the noisy data from comments, we design a weight estimation mechanism which can assign a real value $\rho_i$ for a comment sentence $\mathbf{x}_c^i$. The comment weight $\boldsymbol{\rho} \in \mathbb{R}^{n_c}$ is integrated into the VAEs based sentence modeling and data reconstruction component to handle comments.

\subsection{Reader-Aware Salience Estimation}
Variational Autoencoders (VAEs) \cite{kingma2013auto,rezende2014stochastic} is a generative model based on neural networks which can be used to conduct latent semantic modeling. \citet{li2017salience} employ VAEs to map the news sentences into a latent semantic space, which is helpful in improving the MDS performance. 
Similarly, we also employ VAEs to conduct the semantic modeling for news sentences and comment sentences.
Assume that both the prior and posterior of the latent variables are Gaussian, i.e.,  $p_\theta (\mathbf{z}) = \mathcal{N}(0, \mathbf{I})$ and $q_{\phi}(\mathbf{z}|\mathbf{x}) = \mathcal{N}(\mathbf{z}; \boldsymbol{\mu}, \boldsymbol{\sigma}^2\mathbf{I})$, where  $\boldsymbol{\mu}$ and $\boldsymbol{\sigma}$ denote the variational mean and standard deviation respectively, which can be calculated with a multilayer perceptron (MLP). 
VAEs can be divided into two phases, namely, encoding (inference), and decoding (generation). All the operations are depicted as follows:
\begin{equation}
	\begin{array}{l}
		{h_{enc}} = relu({W_{xh}}x + {b_{xh}})\\
		\mu  = {W_{h\mu }}{h_{enc}} + {b_{h\mu }}\\
		\log ({\sigma ^2}) = {W_{h\sigma }}{h_{enc}} + {b_{h\sigma }}\\
		\varepsilon  \sim \mathcal{N}(0, \mathbf{I}), \ \
		z = \mu  + \sigma  \otimes \varepsilon \\
		{h_{dec}} = relu({W_{zh}}z + {b_{zh}})\\
		{x^\prime} = sigmoid({W_{hx}}{h_{dec}} + {b_{hx}})\\
	\end{array}
	\label{eq:vaes}
\end{equation}

Based on the reparameterization trick in Equation~\ref{eq:vaes},  we can get the analytical representation of the variational lower bound $\mathcal{L}(\theta ,\varphi ;\mathbf{x})$:
\begin{equation}
	\small
	\begin{array}{l}
		\log p(x|z) = \sum\limits_{i = 1}^{|V|} {{x_i}\log x_i^\prime + (1 - {x_i}) \cdot \log (1 - x_i^\prime)} \\
		- {D_{KL}}[{q_\varphi }(z|x)\|{p_\theta }(z)]{\rm{ = }}\frac{1}{2}\sum\limits_{i = 1}^K {(1 + \log (\sigma _i^2) - \mu _i^2 - \sigma _i^2)}
	\end{array}
	\nonumber
\end{equation}
where $\mathbf{x}$ denotes a general sentence, and it can be a news sentence $\mathbf{x}_d$ or a comment sentnece $\mathbf{x}_c$.

By feeding both the news documents and the reader comments into VAEs, we equip  the model a ability of capturing the information from them jointly.
However, there is a large amount of noisy information hidden in the comments. Hence we design a weighted combination mechanism for fusing news and comments in the VAEs. Precisely, we split the variational lower bound $\mathcal{L}(\theta ,\varphi ;\mathbf{x})$ into two parts and fuse them using the comment weight $\boldsymbol{\rho}$:
\begin{equation}
	\mathcal{L}(\theta ,\varphi ;\mathbf{x}) = \mathcal{L}(\theta ,\varphi ;\mathbf{x}_d) + \boldsymbol{\rho} \times \mathcal{L}(\theta ,\varphi ;\mathbf{x}_c)
\end{equation}
The calculation of $\boldsymbol{\rho}$ will be discussed later.

The news sentence salience estimation is conducted by an unsupervised data reconstruction framework.
Assume that $\mathbf{S}_z = \{\mathbf{s}_z^1, \mathbf{s}_z^2,\cdots,\mathbf{s}_z^m\}$ are $m$ latent aspect vectors used for reconstructing all the latent semantic vectors $\mathbf{Z} = \{\mathbf{z}^1, \mathbf{z}^2,\cdots,\mathbf{z}^n\}$. Thereafter, the variational-decoding progress of VAEs can map the latent aspect vector $\mathbf{S}_z$ to $\mathbf{S}_h$, and then produce $m$ new aspect term vectors $\mathbf{S}_x$:
\begin{equation}
	\begin{array}{l}
		{s_{h}} = relu({W_{zh}}{s_z} + {b_{zh}})\\
		{s_x} = sigmoid({W_{hx}}{s_{h}} + {b_{hx}})
	\end{array}
\end{equation}

VAESum \cite{li2017salience} employs an alignment mechanism \cite{bahdanau2014neural,luong2015effective} to recall the lost detailed information from the input sentence.
Inspired this idea, we design a jointly weighted alignment mechanism by considering the news sentence and the comment sentence simultaneously.
For each decoder hidden state $s^i_{h}$, we align it with each news encoder hidden state $h^j_{d}$ by an alignment vector $a^d \in \mathbb{R}^{n_d}$. We also align it with each comments encoder hidden state $h^j_{c}$ by an alignment vector $a^c \in \mathbb{R}^{n_c}$. In order to filter the noisy information from the comments, we again employ the comment weight $\boldsymbol{\rho}$ to adjust the alignment vector of comments:
\begin{equation}
	{\tilde a}^c = a^c \times \boldsymbol{\rho}
\end{equation}

The news-based context vector $c_d^i$ and the comment-based context vector $c_c^i$ can be obtained by linearly blending the input hidden states respectively.
Then the output hidden state can be updated based on the context vectors:
\begin{equation}
	{{\tilde s}_h^i} = \tanh (W_{dh}^hc_d^i + W_{ch}^hc_c^i + W_{hh}^a{s_h^i})
\end{equation}
Then we can generate the updated output aspect vectors based on ${{\tilde s}_h^i}$. We add a similar alignment mechanism into the output layer.

$\mathbf{S}_z$, $\mathbf{S}_h$, and $\mathbf{S}_x$ can be used to reconstruct the space to which they belong respectively.
In order to capture the information from comments, we design a joint reconstruction approach here. 
Let $\mathbf{A}_d \in \mathbb{R}^{n_d \times m}$ be the reconstruction coefficient matrix for news sentences, and $\mathbf{A}_c \in \mathbb{R}^{n_c \times m}$ be the reconstruction coefficient matrix for comment sentences. 
The optimization objective contains three reconstruction terms, jointly considering the latent semantic reconstruction and the term vector space reconstruction for news and comments respectively:
\begin{eqnarray}
	\begin{aligned}
		\mathcal{L}_A &= (\left\| {{Z_d} - {A_d}{S_z}} \right\|_2^2 + \left\| {{H_d} - {A_d}{S_h}} \right\|_2^2 \\
		&+ \left\| {{X_d} - {A_d}{S_x}} \right\|_2^2) + \boldsymbol{\rho} \times (\left\| {{Z_c} - {A_c}{S_z}} \right\|_2^2 \\
		&+ \left\| {{H_c} - {A_c}{S_h}} \right\|_2^2 + \left\| {{X_c} - {A_c}{S_x}} \right\|_2^2)
	\end{aligned}
\end{eqnarray}
This objective is integrated with the variational lower bound of VAEs $\mathcal{L}(\theta ,\varphi ;\mathbf{x})$ and optimized in a multi-task learning fashion. 
Then the new optimization objective is:
\begin{equation}
	\mathcal{J} = \mathop {\min }\limits_\Theta ( {\rm{ - }}\mathcal{L}(\theta ,\varphi ;x){\rm{ + }} {\rm{\mathcal{L}_{A}}})
\end{equation}
where $\Theta$ is a set of all the parameters related to this task.
We define the magnitude of each row of $\mathbf{A}_d$ as the salience scores for the corresponding news sentences.

We should note that the most important variable in our framework is the comment weight vector $\boldsymbol{\rho}$, which appears in all the three components of our framework.
The basic idea for calculating $\boldsymbol{\rho}$ is that if the comment sentence is more similar to the news content, then it contains less noisy information. For all the news sentences $X_d$ and all the comment sentences $X_c$, calculate the relation matrix $R \in \mathbb{R}^{n_d \times n_c}$ by:
\begin{equation}
	R = X_d\times X_c^T
\end{equation}
Then we add an average pooling layer to get the coefficient value for each comment sentence:
\begin{equation}
	\mathbf{r} = \frac{1}{{{n_c}}}\sum\limits_{i = 1}^{{n_c}} {R[i,:]}
\end{equation}
Finally, we add a sigmoid function to adjust the coefficient value to $(0,1)$:
\begin{equation}
	\boldsymbol{\rho} = sigmoid(\mathbf{r})
\end{equation}

Because we have different representations from different vector space for the sentences, therefore we can calculate the comment  weight in different semantic vector space. Here we use two spaces, namely, latent semantic space obtained by VAEs, and the original bag-of-words vector space.
Then we can merge the weights by a parameter $\lambda_p$:
\begin{equation}
	\boldsymbol{\rho} = \lambda_p \times \boldsymbol{\rho}_z + (1-\lambda_p) \times \boldsymbol{\rho}_x
\end{equation}
where $\boldsymbol{\rho}_z$ and $\boldsymbol{\rho}_x$ are the comment weight calculated from latent semantic space and term vector space. Actually, we can regard $\boldsymbol{\rho}$ as some gates to control the proportion of each comment sentence absorbed by the framework.

\subsection{Summary Construction}
\label{sec:ILP_framework}
In order to produce reader-aware summaries, inspired by the phrase-based model in \citet{lidong15absmds} and \citet{li2015reader},
we refine this model to consider the news sentences salience information obtained by our framework.
Based on the parsed constituency tree for each input sentence, we extract the noun-phrases (NPs) and verb-phrases (VPs).
The overall objective function of this optimization formulation for selecting salient NPs and VPs is formulated as an integer linear programming (ILP) problem:
\begin{equation}
	\label{e:objective}
	\max\{\sum_i{\alpha_i S_i} - \sum_{i<j}{\alpha_{ij}(S_i+S_j)R_{ij}}\},
\end{equation}
where $\alpha_i$ is the selection indicator for the phrase $P_i$,
$S_i$ is the salience scores of $P_i$, $\alpha_{ij}$  and $R_{ij}$ is co-occurrence indicator
and the similarity a pair of phrases ($P_i$, $P_j$) respectively.
The similarity is calculated with the Jaccard Index based method.
In order to obtain coherent summaries with good readability,
we add some constraints into the ILP framework. 
For details, please refer to \citet{woodsend2012multiple}, \citet{lidong15absmds}, and \citet{li2015reader}.
The objective function and constraints are linear. Therefore the optimization can be solved by existing ILP solvers such as simplex algorithms~\cite{dantzig2006linear}.
In the implementation, we use a package called lp\_solve\footnote{http://lpsolve.sourceforge.net/5.5/}.

\section{Data Description}
In this section, we describe the preparation process of the dataset.
Then we provide some properties and statistics. 

\subsection{Background}
The definition of the terminology related to the dataset is given as follows.\footnote{In fact, for the core terminology, namely, topic, document, category, and aspect, we follow the MDS task in TAC (\url{https://tac.nist.gov//2011/Summarization/Guided-Summ.2011.guidelines.html}).}\\
\textbf{Topic}: A topic refers to an event and it is composed of a set of news documents from different sources. \\
\textbf{Document}: A news article describing some aspects of the topic. The set of documents in the same topic typically span a period, say a few days.\\
\textbf{Category}: Each topic belongs to a category. There are 6 predefined categories: (1) Accidents and Natural Disasters, (2) Attacks (Criminal/Terrorist), (3) New Technology, (4) Health and Safety, (5) Endangered Resources, and (6) Investigations and Trials (Criminal/Legal/Other).\\
\textbf{Aspect}: Each category has a set of predefined aspects. Each aspect describes one important element of an event. For example, for the category ``Accidents and Natural Disasters'', the aspects are ``WHAT'', ``WHEN'', ``WHERE'', ``WHY'', ``WHO\_AFFECTED'', ``DAMAGES'', and ``COUNTERMEASURES''.\\
\textbf{Aspect facet}: An aspect facet refers to the actual content of a particular aspect for a particular topic. Take the topic ``Malaysia Airlines Disappearance'' as an example, facets for the aspect ``WHAT'' include ``missing Malaysia Airlines Flight 370'', ``two passengers used passports stolen in Thailand from an Austrian and an Italian.'' etc. Facets for the aspect ``WHEN'' are `` Saturday morning'', ``about an hour into its flight from Kuala Lumpur'', etc. \\
\textbf{Comment}: A piece of text written by a reader conveying his or her altitude, emotion, or any thought on a particular news document.

\subsection{Data Collection}

The first step is to select topics. The selected topics should be in one of the above categories. We make use of several ways to find topics. The first way is to search the category name using Google News. The second way is to follow the related tags on Twitter. One more useful method is to scan the list of event archives on the Web, such as earthquakes happened in 2017 \footnote{https://en.wikipedia.org/wiki/Category:2017\_earthquakes}.

For some news websites, in addition to provide news articles, they offer a platform to allow readers to enter comments. Regarding the collection of news documents, for a particular topic, one consideration is that reader comments can be easily found. Another consideration is that all the news documents under a topic must be collected from different websites as far as possible.
Similar to the methods used in DUC and TAC, we also capture and store the content using XML format.

Each topic is assigned to 4 experts, who are major in journalism, to conduct the summary writing.
The task of summary writing is divided into two phases, namely, aspect facet identification, and summary generation.
For the aspect facet identification, the experts read and digested all the news documents and reader comments under the topic.
Then for each aspect, the experts extracted the related facets from the news document.
The summaries were generated based on the annotated aspect facets.
When selecting facets, one consideration is those facets that are popular in both news documents and reader comments have higher priority. Next, the facets that are popular in news documents have the next priority.
The generated summary should cover as many aspects as possible, and should be well-organized using complete sentences with a length restriction of 100 words. 

After finishing the summary writing procedure, we employed another expert for scrutinizing the summaries. Each summary is checked from five
linguistic quality perspectives: grammaticality, non-redundancy, referential clarity, focus, and coherence. Finally, all the model summaries are stored in XML files.

\subsection{Data Properties}
\label{sec:datap}
The dataset contains 45 topics from those 6 predefined categories.
Some examples of topics are ``Malaysia Airlines Disappearance'', ``Flappy Bird'', ``Bitcoin Mt. Gox'', etc. All the topics and categories are listed in Appendix~\ref{sec:apdx_a}.
Each topic contains 10 news documents and 4 model summaries.
The length limit of the model summary is 100 words (slitted by space).
On average, each topic contains 215 pieces of comments and 940 comment sentences.
Each news document contains an average of 27 sentences, and each sentence contains an average of 25 words. 85\% of non-stop model summary terms (entities, unigrams, bigrams) appeared in the news documents, and 51\% of that appeared in the reader comments.
The dataset contains 19k annotated aspect facets.

\section{Experimental Setup}
\subsection{Dataset and Metrics}
The properties of our own dataset are depicted in Section~\ref{sec:datap}.
We use ROUGE score as our evaluation metric \cite{lin2004rouge} with standard options\footnote{ROUGE-1.5.5.pl -n 4 -w 1.2 -m -2 4 -u -c 95 -r 1000 -f A -p 0.5 -t 0}.
F-measures of ROUGE-1, ROUGE-2 and ROUGE-SU4 are reported.

\subsection{Comparative Methods}
To evaluate the performance of our dataset and the proposed framework \textbf{RAVAESum} for RA-MDS, we compare our model with the following methods:

\begin{itemize}
	
	\item \textbf{RA-Sparse} \cite{li2015reader}:  It is a framework to tackle the RA-MDS problem. A sparse-coding-based method is used to calculate the salience of the news sentences by jointly considering news documents and reader comments.
	
	\item \textbf{Lead} \cite{wasson1998using} : It ranks the news sentences chronologically and extracts the leading sentences one by one until the length limit.

	\item \textbf{Centroid} \cite{radev2000centroid}: It summarizes clusters of news articles automatically grouped by a topic detection system, and then it uses information from the centroids of the clusters to select sentences. 
	
	\item \textbf{LexRank} \cite{erkan2004lexrank} and \textbf{TextRank} \cite{mihalcea2004textrank}: Both methods are graph-based unsupervised framework for sentence salience estimation based on PageRank algorithm. 
	
	\item \textbf{Concept} \cite{lidong15absmds}: It generates abstractive summaries using phrase-based optimization framework with concept weight as salience estimation. The concept set contains unigrams, bigrams, and entities. The weighted term-frequency is used as the concept weight.
	
\end{itemize}
We can see that only the method RA-Sparse can handle RA-MDS.
All the other methods are only for traditional MDS without comments.

\subsection{Experimental Settings}
The input news sentences and comment sentences are represented as BoWs vectors with dimension $|V|$.
The dictionary $V$ is created using unigrams, bigrams and named entity terms. $n_d$ and $n_c$ are the number of news sentences and comment sentences respectively. For the number of latent aspects used in data reconstruction, we let $m = 5$.
For the neural network framework, we set the hidden size $d_h = 500$ and the latent size $K = 100$. For the parameter $\lambda_p$ used in comment weight, we let $\lambda_p=0.2$. 
Adam \cite{kingma2014adam} is used for gradient based optimization with a learning rate 0.001.
Our neural network based framework is implemented using Theano \cite{bastien2012theano} on a single GPU\footnote{Tesla K80, 1 Kepler GK210 is used, 2496 Cuda cores, 12G GDDR5 memory.}. 

\section{Results and Discussions}

\subsection{Results on Our Dataset}

The results of our framework as well as the baseline methods are depicted in Table~\ref{tab:rouge}.
It is obvious that our framework RAVAESum is the best among all the comparison methods.
Specifically, it is better than RA-Sparse significantly ($p<0.05$), which demonstrates that VAEs based latent semantic modeling and joint semantic space reconstruction can improve the MDS performance considerably. 
Both RAVAESum and RA-Sparse are better than the methods without considering reader comments.

\begin{table}[!t]
	\begin{threeparttable}
		\centering
		\caption{Summarization performance.}
		\label{tab:rouge}
		\begin{tabular}{p{2.2cm} c c c}
			\hline
			\textbf{System}  & \textbf{R-1} & \textbf{R-2} & \textbf{R-SU4} \\
			\hline
			Lead           & 0.384 & 0.110  & 0.144 \\
			TextRank       & 0.402 & 0.122  & 0.159 \\
			LexRank        & 0.425 & 0.135  & 0.165 \\
			Centroid       & 0.402 & 0.141  & 0.171 \\
			Concept        & 0.422 & 0.149  & 0.177 \\
			RA-Sparse         & 0.442 & 0.157  & 0.188 \\
			\hline
			RAVAESum        &  \ \ \textbf{0.443*} &  \ \ \textbf{0.171*} & \ \ \textbf{0.196*} \\			
			\hline
		\end{tabular}
	\end{threeparttable}
\end{table}

\subsection{Further Investigation of Our Framework }

\begin{table}[!t]
	\begin{threeparttable}
		\centering
		\caption{Further investigation of RAVAESum.}
		\label{tab:eff_cmt}
		\begin{tabular}{p{2.2cm} c c c}
			\hline
			\textbf{System}  & \textbf{R-1} & \textbf{R-2} & \textbf{R-SU4} \\
			\hline
			RAVAESum-noC      & 0.437 & 0.162  & 0.189 \\
			\hline
			RAVAESum        &  \ \ \textbf{0.443*} &  \ \ \textbf{0.171*} & \ \ \textbf{0.196*} \\			
			\hline
		\end{tabular}
	\end{threeparttable}
\end{table}

To further investigate the effectiveness of our proposed RAVAESum framework, we adjust our framework by removing the comments related components. Then the model settings of RAVAESum-noC are similar to VAESum \cite{li2017salience}.
The evaluation results are shown in Table~\ref{tab:eff_cmt}, which illustrate that our framework with reader comments RAVAESum is better than RAVAESum-noC significantly($p<0.05$). 

\begin{table*}[!t]
	\centering
	\caption{Top-10 terms extracted from each topic according to the word salience values}
	\label{tab:wrod-salience}
	\begin{tabular}{|l|l|l|} \hline
		\textbf{\ \ \ \ \ \ Topic} &\textbf{$\pm$C}& \textbf{\ \ \ \ \ \ \ \ \ \ \ \ \ \ \ \ \ \ \ \ \ \ \  \ \ \ \ \ \ \ \ \ \ \ \ \ \ \ \ \ \ \ \ \ \ Top-10 Terms}\\
		\hline
		\multirow{2}{2.2cm}{``Sony Virtual Reality PS4''}
		& $-$C &Sony, headset, game, virtual, morpheus, reality, vr, project, playstation, Yoshida\\
		& $+$C &Sony, game, vr, virtual, headset, reality, morpheus, \textbf{\textit{oculus}}, project, playstation\\
		\hline
		\multirow{2}{2.2cm}{``Bitcoin Mt. Gox Offlile''} & $-$C &bitcoin, gox, exchange, mt., currency, Gox, virtual, company, money, price\\
		& $+$C &bitcoin, currency, money, exchange, gox, mt., virtual, company, price, world\\
		\hline 
	\end{tabular}
\end{table*}

Moreover, as mentioned in VAESum \cite{li2017salience}, the output aspect vectors contain the word salience information.
Then we select the top-10 terms for event ``Sony Virtual Reality PS4'', and ```Bitcoin Mt. Gox Offlile''' for model RAVAESum (+C) and RAVAESum-noC (-C) respectively, and the results are shown in Table~\ref{tab:wrod-salience}.
It is obvious that the rank of the top salience terms are different.
We check from the news documents and reader comments and find that some terms are enhanced by the reader comments successfully.
For example, for the topic ``Sony Virtual Reality PS4'', many readers talked about the product of ``Oculus'', hence the word ``oculus'' is assigned a high salience by our model.

\subsection{Case Study}

\begin{table}[!t]
	\centering
	\caption{Generated summaries for the topic ``Sony Virtual Reality PS4''.}
	\begin{tabular}{|p{2.8cm} | c| c| c|}
		\hline
		\textbf{System}  & \textbf{R-1} & \textbf{R-2} & \textbf{R-SU4} \\
		\hline
		RAVAESum-noC & 0.482 & 0.184 & 0.209 \\
		\hline
		\multicolumn{4}{|p{7.2cm}|}{
			\footnotesize
			A virtual reality headset that's coming to the PlayStation 4.
			\textbf{\textit{Today announced the development of ``Project Morpheus'' (Morpheus) ''a virtual reality (VR) system that takes the PlayStation4 (PS4)''.}}
			Shuhei Yoshida, president of Sony Computer Entertainment, revealed a prototype of Morpheus at the Game Developers Conference in San Francisco on Tuesday.
			Sony showed off a prototype device V called Project Morpheus V that can be worn to create a virtual reality experience when playing games on its new PlayStation 4 console.
			\textbf{\textit{The camera on the Playstation 4 using sensors that track the player's head movements.}}}\\
		\hline
		RAVAESum       & \textbf{0.490} & \textbf{0.230} & \textbf{0.243}\\
		\hline
		\multicolumn{4}{|p{7.2cm}|}{
			\footnotesize
			Shuhei Yoshida, president of Sony Computer Entertainment, revealed a prototype of Morpheus at the Game Developers Conference in San Francisco on Tuesday.
			A virtual reality headset that's coming to the PlayStation 4.
			Sony showed off a prototype device V called Project Morpheus V that can be worn to create a virtual reality experience when playing games on its new PlayStation 4 console.
			\textbf{\textit{Mr. Yoshida said that Sony was inspired and encouraged to do its own virtual reality project after the enthusiastic response to the efforts of Oculus VR and Valve, another game company working on the technology.}}}\\
		\hline
	\end{tabular}
	\label{tab:case1}
\end{table}

Based on the news and comments of the topic ``Sony Virtual Reality PS4'', we generate two summaries with our model considering comments (RAVAESum) and ignoring comments (RAVAESum-noC) respectively.
The summaries and ROUGE evaluation are given in Table \ref{tab:case1}. All the ROUGE values of our model considering comments are better than those ignoring comments with large gaps.
The sentences in \emph{\textbf{italic bold}} of the two summaries are different. By reviewing the comments of this topic, we find that many readers talked about ``Oculus'', the other product with virtual reality techniques. This issue is well identified by our model and select the sentence ``\textit{Mr. Yoshida said that Sony was inspired and encouraged to do its own virtual reality project after the enthusiastic response to the efforts of Oculus VR and Valve, another game company working on the technology.}''.

\section{Conclusions}
We investigate the problem of reader-aware multi-document summarization (RA-MDS) and introduce a new dataset. To tackle the RA-MDS, we extend a variational auto-encodes (VAEs) based MDS framework by jointly considering news documents and reader comments. The methods for data collection, aspect annotation, and summary writing and scrutinizing by experts are described. Experimental results show that reader comments can improve the summarization performance, which demonstrate the usefulness of the proposed dataset.

\bibliography{emnlp2017}
\bibliographystyle{emnlp_natbib}

\clearpage
\begin{appendices}
	
	\section{Topics}
	\label{sec:apdx_a}
	
	\begin{table}[H]
		\centering
		\caption{All the topics and the corresponding categories. The 6 predefined categories are: (1) Accidents and Natural Disasters, (2) Attacks (Criminal/Terrorist), (3) New Technology, (4) Health and Safety, (5) Endangered Resources, and (6) Investigations and Trials (Criminal/Legal/Other).}
		\label{tab:alltopics}
		\small
		\begin{tabular}{p{5.8cm}|c}
			\hline
			\textbf{Topic} & \textbf{Category}\\
			\hline
			Boston Marathon Bomber Sister Arrested & 6 \\ 
			\hline
			iWatch & 3 \\ 
			\hline
			Facebook Offers App With Free Access in Zambia & 3 \\ 
			\hline
			441 Species Discovered in Amazon & 5 \\ 
			\hline
			Beirut attack & 2 \\ 
			\hline
			Great White Shark Choked by Sea Lion & 1 \\ 
			\hline
			Sony virtual reality PS4 & 3 \\ 
			\hline
			Akademik Shokalskiy Trapping & 1 \\ 
			\hline
			Missing Oregon Woman Jennifer Huston Committed Suicide & 6 \\ 
			\hline
			Bremerton Teen Arrested Murder 6-year-old Girl & 6 \\ 
			\hline
			Apple And IBM Team Up & 3 \\ 
			\hline
			California Father Accused Killing Family & 6 \\ 
			\hline
			Los Angeles Earthquake & 1 \\ 
			\hline
			New Species of Colorful Monkey & 5 \\ 
			\hline
			Japan Whaling & 5 \\ 
			\hline
			Top Doctor Becomes Latest Ebola Victim & 4 \\ 
			\hline
			New South Wales Bushfires & 1 \\ 
			\hline
			UK David Cameron Joins Battle Against Dementia & 4 \\ 
			\hline
			UK Cameron Calls for Global Action on Superbug Threat & 4 \\ 
			\hline
			Karachi Airport Attack & 2 \\ 
			\hline
			Air Algerie Plane Crash & 1 \\ 
			\hline
			Flappy Bird & 3 \\ 
			\hline
			Moscow Subway Crash & 1 \\ 
			\hline
			Rick Perry Lawyers Dismissal of Charges & 6 \\ 
			\hline
			New York Two Missing Amish Girls Found & 6 \\ 
			\hline
			UK Contaminated Drip Poisoned Babies & 4 \\ 
			\hline
			Taiwan Police Evict Student Protesters & 2 \\ 
			\hline
			US General Killed in Afghan & 5 \\ 
			\hline
			Monarch butterflies drop & 5 \\ 
			\hline
			UN Host Summit to End Child Brides & 4 \\ 
			\hline
			Two Tornadoes in Nebraska & 1 \\ 
			\hline
			Global Warming Threatens Emperor Penguins & 5 \\ 
			\hline
			Malaysia Airlines Disappearance & 1 \\ 
			\hline
			Google Conference & 3 \\ 
			\hline
			Africa Ebola Out of Control in West Africa & 4 \\ 
			\hline
			Shut Down of Malaysia Airlines mh17 & 1 \\ 
			\hline
			Sochi Terrorist Attack & 2 \\ 
			\hline
			Fire Phone & 3 \\ 
			\hline
			ISIS executes David Haines & 2 \\ 
			\hline
			UK Rotherham 1400 Child Abuse Cases & 6 \\ 
			\hline
			Rare Pangolins Asians eating Extinction & 5 \\ 
			\hline
			Kunming Station Massacre & 2 \\ 
			\hline
			Bitcoin Mt. Gox & 3 \\ 
			\hline
			UK Jimmy Savile Abused Victims in Hospital & 6 \\ 
			\hline
			ISIS in Iraq & 2 \\ 
			\hline
		\end{tabular}
	\end{table}

\end{appendices}

\end{document}